\documentclass[a4paper, 12pt]{article}
\usepackage[english]{babel}
\usepackage[utf8]{inputenc}
\usepackage[centertags]{amsmath}
\usepackage{amssymb,amsfonts,amsthm}
\usepackage{indentfirst}
\usepackage{array}
\usepackage{float}
\usepackage[usenames,dvipsnames,svgnames,table]{xcolor}
\usepackage{hyperref}
\usepackage{enumerate}
\usepackage{graphicx}
\usepackage{epstopdf}
\usepackage{graphics}
\usepackage{subfigure}
\usepackage[margin=1.5cm,footskip=1.5cm,includefoot]{geometry} 
\usepackage{listings}
\usepackage{setspace}
\usepackage{multirow}
\usepackage{lipsum}
\usepackage[utf8]{inputenc}
\usepackage{dsfont,amsfonts}
\usepackage{graphicx}
\usepackage{indentfirst}
\usepackage{subfigure}
\usepackage{listings}
\everymath{\displaystyle}
\usepackage{amsmath}
\usepackage{booktabs}

\usepackage{tikz}
\usetikzlibrary{matrix}

\newcommand{\bx}{{\bf x}}
\newcommand{\bw}{{\bf w}}
\newcommand{\bm}{{\bf m}}
\newcommand{\ba}{{\bf a}}
\newcommand{\bb}{{\bf b}}

\newtheorem{theorem}{Theorem}[section]
\newtheorem{example}[theorem]{Example}

\begin{document}
	\begin{center}
		\sc\large\textbf{XV Encontro Científico de Pós-Graduandos do IMECC}\\
		\vspace{-1em}
		\noindent \hrulefill\\	
		\vspace{-1.1em}
		\noindent \hrulefill\\
	\end{center}
	
	\vspace{-.7cm}
	\begin{center}
		\Large{Linear Dilation-Erosion Perceptron for Binary Classification}
	\end{center}
	\vspace{-.1cm}
	
	\noindent {\small 
	\underline{Angelica Lourenço Oliveira}, 
	Marcos Eduardo Valle. ra211686@ime.unicamp.br, valle@ime.unicamp.br\\}
	{\footnotesize Department of Applied Mathematics, IMECC, UNICAMP, Brazil.}
	
	\begin{abstract} In this work, we briefly revise the reduced dilation-erosion perceptron (r-DEP) models for binary classification tasks. Then, we present the so-called linear dilation-erosion perceptron (l-DEP), in which a linear transformation is applied before the application of the morphological operators. Furthermore, we propose to train the l-DEP classifier by minimizing a regularized hinge-loss function subject to concave-convex restrictions. A simple example is given for illustrative purposes.

	\noindent {\bf Keys words}:  {morphological neural networks, dilation-erosion perceptron, concave-convex programming. }
	\vspace{-.5cm}
	\end{abstract}
	
	\section{Reduced Dilation-Erosion Perceptron}
	\vspace{-.3cm}
	Morphological neural networks (MNNs) are machine learning models whose processing units perform operations from mathematical morphology \cite{sussner_morphological_2011}. 
	
	In \cite{de_a_araujo_class_2011}, Araujo proposed a hybrid MNN called dilation-erosion perceptron (DEP). For binary classificaiton tasks, a DEP computes the convex combination of a dilation and an erosion followed by a hard-limiter function. Formally, a DEP model is given by ${\phi(\bx)=\mbox{sgn}(\beta\delta_{\ba}(\bx)+(1-\beta) \varepsilon_{\bb}(\bx))}$, for some $\beta\in [0,1]$.	Using concepts from tropical algebra, Charisopoulos and Maragos formulated the training of MNNs as well as the hybrid DEP model as the solution of a convex-concave programming (CCP) problem  \cite{charisopoulos_morphological_2017}. 
	
    Despite the encouraging results obtained using CCP training in some well-known classification tasks, the DEP model assumes an ordering relationship between features and classes, that is, samples from one class should be less than or equal to the samples from the other class. Taking into account reduced orderings widely used in multivariate mathematical morphology, Valle recently proposed the so-called reduced DEP (r-DEP) which overcomes the limitations of the DEP classifier \cite{valle_reduced_2020}. However, an effective r-DEP model needs an appropriate surjective mapping $\rho$ which transforms the feature space into $\mathds{R}^n$. The DEP model is then applied to the transformed data, that is, a r-DEP model is given by the composition $\phi^r = \phi \circ \rho$.  
	\vspace{-.4cm}
	\section{Linear Dilation-Erosion Perceptron}
	\vspace{-.3cm}

    How to determine the surjective mapping $\rho$ is one of the main challenges in defining a successful r-DEP classifier. In a linear dilation-erosion perceptron (l-DEP), the mapping $\rho$ is linear. The application of linear transformations before the evaluation of an elementary morphological operator yields the classifier 
    $\phi^l(\bx) = \mbox{sgn} \left( \beta \delta_{\ba}(W \bx) + (1-\beta) \varepsilon_{\bb}(M \bx)\right)$,
	where $W \in \mathds{R}^{n_1 \times n}$ and $M \in \mathds{R}^{n_2 \times n}$. Redefining the parameters by incorporating $\beta$ into $W$ and $\ba$, and $(\beta-1)$ into $M$ and $\bb$, the l-DEP model can be alternatively given by $\phi^l(\bx) = \mbox{sgn}(\tau\big(\bx)\big)$, with $\tau(\bx) = \max_{i=1:n_1}(\bw_i^T \bx + a_i) - \max_{j=1:n_2}(\bm_j^T \bx + b_j),$ 	where $\bw_i^T$ and $\bm_j^T$ are rows of $W$ and $M$, respectively. The function $\tau$ is the decision function of the l-DEP classifier.
	
	Given a training set $T=\{(\bx_i,d_i):i=1,\ldots,m \} \subseteq \mathds{R}^n \times \{-1,+1\}$, the parameters $W$, $\ba$, $M$ and $\bb$ of a l-DEP classifier can be determined by solving a disciplined convex-concave programming problem (DCCP) as follows. Let $C^+ = \{\bx_i:d_i = +1\}$ and $C^- = \{\bx_i:d_i = -1\}$ be the sets of feature samples from the positive and negative classes, respectively. Inspired by the linear support vector machine and elastic net, we propose to train the l-DEP model by solving the DCCP problem
	\vspace{-.3cm}
    \begin{equation} \label{eq:DCCP}
        \begin{cases}
        \mathop{\mbox{minimize}}_{W,\ba,M,\bb,\boldsymbol{\xi}} & \frac{C}{m}\sum_{i=1}^m \max\{\xi_i,0\}+ r_W + r_M, \\ 
          \mbox{subject to} & \max_{i=1:n_1}(\bw_i^T \bx + a_i) + 1 \leq \max_{j=1:n_2}(\bm_j^T \bx + b_j) + \xi_i, \quad \forall \bx_i \in C^-, \\
          & \max_{i=1:n_1}(\bw_i^T \bx + a_i) + \xi_i \geq \max_{j=1:n_2}(\bm_j^T \bx + b_j) + 1, \quad \forall \bx_i \in C^+,
        \end{cases}
    \end{equation}
    where ${r_W = \lambda_W((1-\alpha)\|W\|_F^2 + \alpha\sum_{j}\|\bw_j\|_1 )}$ and ${r_M=\lambda_M((1-\alpha)\|M\|_F^2 + \alpha\sum_{j}\|\bm_j\|_1 )}$ are elastic net regularizations, and $\alpha \in [0,1]$, $\lambda_W, \lambda_M,C\in \mathds{R}$ are normalizing constants. The slack variables $\xi_i$, for $i=1,\ldots,m$, allow classification errors. The terms $+1$ in the constrains maximize the margin of separation between the positive and negative training samples. Note that the parameter $W,M,\ba ,\bb,\boldsymbol{\xi}$ are self-adjusting in the training of the l-DEP model. In this work, we solved the DCCP problem \eqref{eq:DCCP} using the algorithm proposed by Shen et al. \cite{shen_disciplined_2016}. In our computational implementations and experiments, we use the \texttt{CVXPY} package, which has an extension for solving \texttt{DCCP} problems, combined with the \texttt{MOSEK} solver. Bellow, we provide an example to illustrate the l-DEP model.
    \begin{example}The Ripley dataset consists of a set of synthetic data with $250$ training samples and $1000$ test samples. We adopted the constants $C = 1$, $\alpha = 1$, and $\lambda = 5e^{-4}$ as well as $m_1 = 4$ and $m_2 = 3$. The l-DEP classifier achieved accuracies of $90 \%$ and $90 \%$ in the training and test sets, respectively. The decision boundary of the l-DEP classifier as well as the parameters obtained solving the DCCP problem \eqref{eq:DCCP} are given below:
    
    \vspace{-.6cm}
    \begin{tabular}{cc}
    \parbox{0.5\columnwidth}{
    $W = \left[\begin{tabular}{cc}
    \, 0.000\, & \,-4.456 \\
    \,-6.828\, & \, 5.977 \\
    \, 7.438\, & \, 3.109 \\
    \,-0.000\, & \,-0.000
    \end{tabular}\right], \quad
    \ba = \left[\begin{tabular}{c}
     4.532 \\
     0.148 \\
    -0.829 \\
     1.854
    \end{tabular}\right]$,
    
    $M = \left[\begin{tabular}{cc}
     0.000 & -4.456 \\
    -19.349 & -0.000 \\
    -0.000 & -0.000
    \end{tabular}\right],\quad
    \bb = \left[\begin{tabular}{c}
    -5.532 \\
     2.955 \\
    -1.285
    \end{tabular}\right].$}
    &
    \parbox{0.5\columnwidth}{
    \includegraphics[width=0.338\columnwidth]{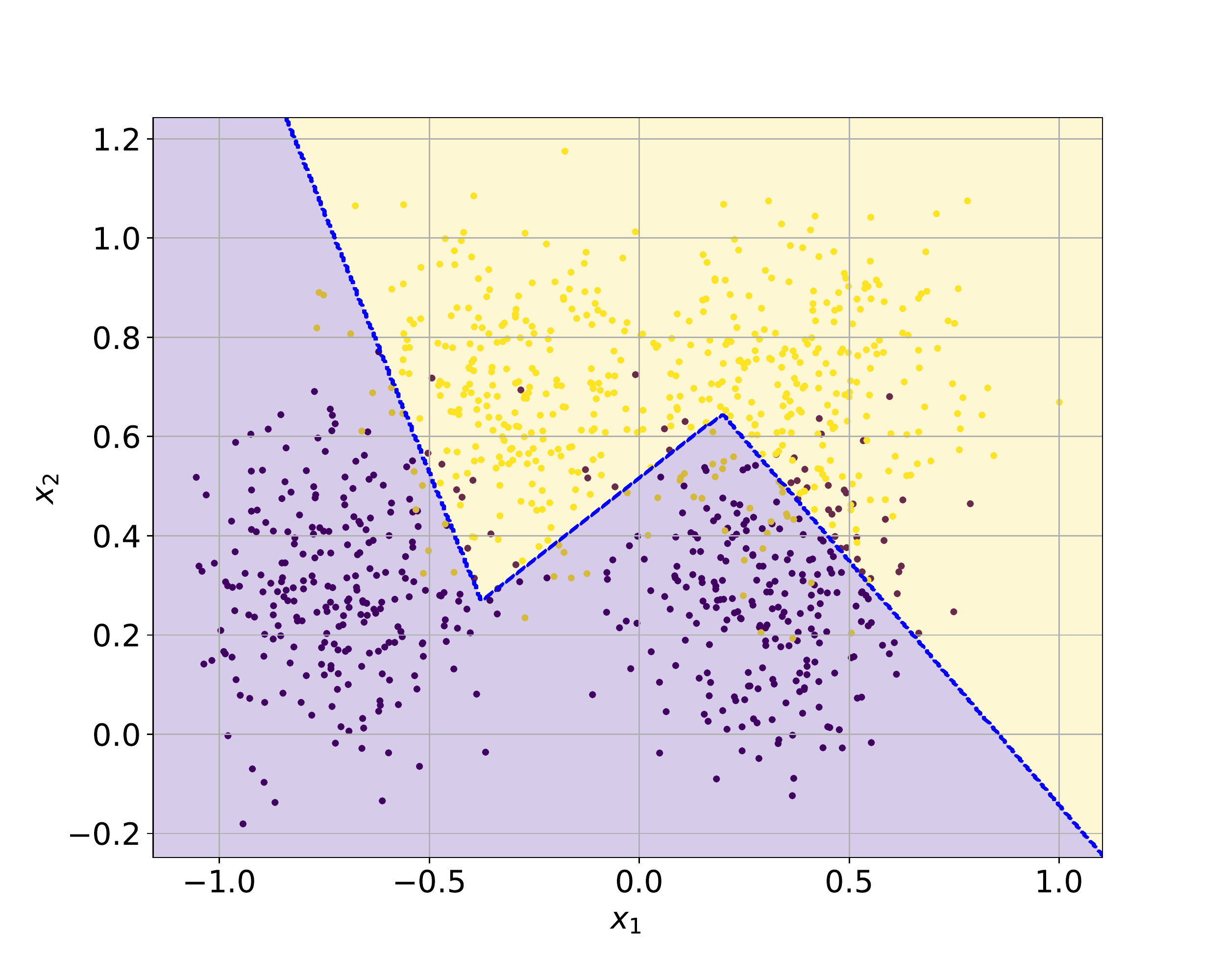} \\
    Figure 1. Decision boundary.}
    \end{tabular}
    \end{example}
    \vspace{-1cm}
    
	{\small 
    
    \vspace{-.1cm}
        }
\end{document}